\documentclass[a4paper,conference]{IEEEtran}

\def\BibTeX{{\rm B\kern-.05em{\sc i\kern-.025em b}\kern-.08emT\kern-.1667em\lower.7ex\hbox{E}\kern-.125emX}}

\usepackage[utf8]{inputenc} 
\usepackage[T1]{fontenc}    
\usepackage{hyperref}       
\usepackage{url}            
\usepackage{booktabs}       
\usepackage{amsfonts}       
\usepackage{nicefrac}       
\usepackage{microtype}      
\usepackage{graphicx}
\usepackage{wrapfig}
\usepackage{adjustbox}
\usepackage{sidecap}
\usepackage{subcaption}
\usepackage[font=small,skip=0pt]{subcaption}
\usepackage{makecell}
\usepackage{color}
\usepackage{amsmath}
\usepackage{wrapfig}
\usepackage{lineno}
\usepackage{enumitem}
\usepackage{comment}
\usepackage{natbib}

\newcommand{\tb}[1]{{\color{magenta}\textbf{TB}: #1}}

\ifCLASSINFOpdf
\else
\fi

\begin{document}
%
\title{Improving Model Accuracy for Imbalanced Image
Classification Tasks by Adding a Final Batch Normalization Layer: An Empirical Study}

\author{\IEEEauthorblockN{Veysel~Kocaman}
\IEEEauthorblockA{LIACS\\ 
  Leiden University\\
  Leiden, The Netherlands\\
v.kocaman@liacs.leidenuniv.nl}
\and
\IEEEauthorblockN{Ofer~M.~Shir}
\IEEEauthorblockA{Computer Science Department\\ 
  Tel-Hai College and Migal Institute,\\
  Upper Galilee, Israel\\
ofersh@telhai.ac.il}
\and
\IEEEauthorblockN{Thomas~Bäck}
\IEEEauthorblockA{LIACS\\ 
  Leiden University\\
  Leiden, The Netherlands\\
t.h.w.baeck@liacs.leidenuniv.nl}}


%


\maketitle

\begin{abstract}
Some real-world domains, such as Agriculture and Healthcare, comprise early-stage disease indications whose recording constitutes a rare event, and yet, whose precise detection at that stage is critical. 
In this type of highly imbalanced classification problems, which encompass complex features, deep learning (DL) is much needed because of its strong detection capabilities. At the same time, DL is observed in practice to favor majority over minority classes and consequently suffer from inaccurate detection of the targeted early-stage indications. 
To simulate such scenarios, we artificially generate skewness (99\% vs.~1\%) for certain plant types out of the PlantVillage dataset as a basis for classification of scarce visual cues through transfer learning. 
By randomly and unevenly picking healthy and unhealthy samples from certain plant types to form a training set, we consider a base experiment as fine-tuning ResNet34 and VGG19 architectures and then testing the model performance on a balanced dataset of healthy and unhealthy images. We empirically observe that the initial F1 test score jumps from 0.29 to 0.95 for the minority class upon adding a final Batch Normalization (BN) layer just before the output layer in VGG19.  We demonstrate that utilizing an additional BN layer before the output layer in modern CNN architectures has a considerable impact in terms of minimizing the training time and testing error for minority classes in highly imbalanced data sets. 
Moreover, when the final BN is employed, minimizing the loss function may not be the best way to assure a high F1 test score for minority classes in such problems. 
That is, the network might perform better even if it is not ‘confident’ enough while making a prediction; leading to another discussion about why softmax output is not a good uncertainty measure for DL models.
We also report on the corroboration of these findings on the ISIC Skin Cancer as well as the Wall Crack datasets.
\end{abstract}


%
\IEEEpeerreviewmaketitle

\section{Introduction}
\label{sec:introduction}
Detecting anomalies that are hardly distinguishable from the majority of observations is a challenging task that often requires strong learning capabilities.
Since anomalies appear scarcely, and in instances of diverse nature, a labeled dataset representative of all forms is typically unattainable. Despite tremendous advances in computer vision and object recognition algorithms, their effectiveness remains strongly dependent upon the size and distribution of the training set. Real-world settings dictate hard limitations on training sets of rarely recorded events in Agriculture or Healthcare, e.g., early stages of certain crop diseases or premature malignant tumors in humans. 
The current work is rooted in Precision Agriculture, and particularly in Precision Crop Protection \cite{mahlein_recent_2012}, where certain visual cues must be recognized with high accuracy in early stages of infectious diseases' development. A renowned use-case is the Potato Late Blight, with dramatic historical and economical impacts \cite{PotatoLateBlightReview2014}, whose early detection in field settings remains an open challenge (despite progress achieved in related learning tasks; see, e.g., \cite{Monsanto2018}). The hard challenge stems from the actual nature of the visual cues (which resemble soil stains and are hardly distinguishable), but primarily from the fact that well-recording those early-stage indications is a rare event. 

In recent years, reliable models capable of learning from small samples have been obtained through various approaches, such as autoencoders~\cite{beggel2019robust}, fine tuning with transfer learning~\cite{hussain2018study}, data augmentation~\cite{shorten2019survey}, cosine loss utilizing (replacing categorical cross entropy)~\cite{barz2020deep}, or prior knowledge~\cite{lake2015human}. 
Our primary research question is thus the following: \textit{What is the most effective approach to enable learning of minority classes, and could Batch Normalization (BN) serve as one?}
\begin {comment}
\begin{quote} \tb { (vk) can we make this sentence italic and shift above next to colon so that we save some space :-)}
    What is the most effective approach to enable learning of minority classes, and could Batch Normalization (BN) serve as one?
\end{quote}
\end{comment}

A prominent effort towards plant diseases' detection is the PlantVillage Disease Classification Challenge. As part of the PlantVillage project, 54,306 images of 14 crop species with 26 diseases (including healthy, 38 classes in total) were made publicly available~\cite{hughes2015open}. 
Following the release of the PlantVillage dataset, deep learning (DL) was intensively employed~\cite{mohanty2016using}, with reported accuracy ranging from 85.53\% to 99.34\%.
Without any feature engineering, the best reported model correctly classifies crop and disease in 993 out of 1,000 images (out of 38 possible classes). The PlantVillage dataset is slightly imbalanced, such that accuracy is commonly used as the performance measure.

In this study, to simulate rare events in agriculture, we capitalize on this dataset to synthetically generate skewness and address our research question.
Upon firstly reproducing the accuracy rates reported in \cite{mohanty2016using} on the original dataset using a ResNet34 architecture~\cite{he2016deep}, we considered the classification problem under the generated imbalance.
We randomly picked 1,000 healthy and 10 unhealthy samples from a plant type for the training set, 150 healthy and 7 unhealthy samples for the validation set, fine tuned using the ResNet34 architecture, and then tested the model performance on an equal number of healthy and unhealthy images (150 vs 150) in a test set. We did this for 3 different plant types: Apple, Pepper, and Tomato. Training on just 10 samples and validating on 7 samples from minority class, we managed to correctly predict more than 141 of 150 unhealthy images spanning over multiple anomalies that may not exist in the training or validation set. When experimenting with different configurations, we discovered a significant improvement of classification performance by adding a final BN layer just before the output layer. We did more experiments in the VGG19 framework~\cite{simonyan2014very} under the same settings, and managed to correctly predict more than 143 of 150 unhealthy images, which is higher than what we got through ResNet34: 130 out of 150.

Altogether, we will show that irrespective of freezing or unfreezing the previous BN layers in the pre-trained modern CNN architectures, putting an additional BN layer just before the softmax output layer has a considerable impact in terms of minimizing the training time and test error for minority classes in a highly imbalanced dataset (especially in a setting where a high recall is desired for the minority class). Our experiments show that using ResNet34 we can achieve an F1 test score of 0.94 in 10 epochs for the minority class while we could only achieve 0.42 without using the final BN layer. Using VGG19, the same approach increases the F1 test score from 0.25 to 0.96 in 10 epochs. We also report that these findings were also corroborated on additional anomaly detection datasets, namely the ISIC Skin Cancer~\cite{gutman2016skin} and the Wall Crack datasets~\cite{zhang2016road}, under the same imbalance settings. 

The concrete contributions of this paper are the following:
\begin{itemize}
\item Putting an additional BN layer just before the output layer has a considerable impact in terms of minimizing the training time and test error for minority classes in highly imbalanced image classification tasks. Our experiments show that the initial F1 test score increases from the 0.29-0.56 range to the 0.95-0.98 range for the minority class when we added a final BN layer just before the softmax output layer. 

\item Trying to minimize validation and train losses may not be an optimal way for getting a high F1 test score for minority classes in binary anomaly detection problems. We illustrate that having a higher training and validation loss but high validation accuracy leads to higher F1 test scores for minority classes in less time and using the final BN layer has a calibration effect. That is, the network might perform better even if it is not ‘confident’ enough while making a prediction. We will also argue that lower values in softmax output may not necessarily indicate ‘lower confidence level’, leading to another discussion about why softmax output is not a good uncertainty measure for deep learning (DL) models.

\end{itemize}

The remainder of the paper is organized as follows: 
Section~\ref{sec:background} gives some background concerning the role of the BN layer in CNN.
Section~\ref{sec:relatedWork} provides an overview of the related work. 
Section~\ref{sec:ImplementationDetails} elaborates the implementation details and settings for our experiments and presents results.
Section~\ref{sec:discussion}  discusses the findings and proposes possible \textit{mechanistic} explanations.
Section~\ref{sec:conclusion} concludes this paper by pointing out key points and future directions.

\section{Background}
\label{sec:background}
In order to better understand the novel contributions of this study, 
in this section, we give some background information about the Batch Normalization~\cite{ioffe2015batch} concept.

Training deep neural networks with dozens of layers is challenging as they can be sensitive to the initial random weights and configuration of the learning algorithm. One possible reason for this difficulty is that the distribution of the inputs to layers deep in the network may change after each mini-batch when the weights are updated. This slows down the training by requiring lower learning rates and careful parameter initialization, makes it notoriously hard to train models with saturating nonlinearities~\cite{ioffe2015batch}, and can cause the learning algorithm to forever chase a moving target. This change in the distribution of inputs to layers in the network is referred to by the technical name “internal covariate shift” (ICS).

BN is a widely adopted technique that is designed to combat ICS and to enable faster and more stable training of deep neural networks (DNNs). It is an operation added to the model before activation which normalizes the inputs and then applies learnable scale ($\gamma$) and shift ($\beta$) parameters to preserve model performance. Given $m$ activation values $x_1\ldots,x_m$ from a mini-batch ${\cal B}$ for any particular layer input $x^{(j)}$ and any dimension $j \in \{1,\ldots,d\}$, the transformation uses the mini-batch mean 
$\mu_{\cal B}= 1/m \sum_{i=1}^m x_i$ and variance $\sigma_{\cal B}^2 = 1/m \sum_{i=1}^m (x_i - \mu_{\cal B})^2$ for normalizing the $x_i$ according to $\hat{x}_i = (x_i - \mu_{\cal B})/\sqrt{\sigma_{\cal B}^2 + \epsilon}$
%
%
and then applies the scale and shift to obtain the transformed values $y_i$ = $\gamma \hat{x}_i + \beta$. The constant $\epsilon > 0$ assures numerical stability of the transformation.


BN has the effect of stabilizing the learning process and dramatically reducing the number of training epochs required to train deep networks; and using BN 
makes the network more stable during training. This may require the use of much larger than normal learning rates, which in turn may further speed up the learning process. 

Though BN has been around for a few years and has become common in deep architectures, it remains one of the DL concepts that is not fully understood, having many studies discussing why and how it works.
Most notably, Santurkar et al.~\cite{santurkar2018does} recently demonstrated that such distributional stability of layer inputs has little to do with the success of BN and the relationship between ICS and BN is tenuous. Instead, they uncovered a more fundamental impact of BN on the training process: it makes the optimization landscape significantly smoother. This smoothness induces a more predictive and stable behavior of the gradients, allowing for faster training. Bjorck et al.~\cite{bjorck2018understanding} also makes similar statements that the success of BN can be explained without ICS. They argue that being able to use larger learning rate increases the implicit regularization of the gradient, which improves generalization. 
 
Even though BN adds an overhead to each iteration (estimated as additional 30\% computation~\cite{mishkin2015all}), the following advantages of BN outweigh the overhead shortcoming:
\begin{itemize}
\item It improves gradient flow 
and allows training deeper models (e.g., ResNet).
\item It enables using higher learning rates because it eliminates outliers activation, hence the learning process may be accelerated using those high rates.
\item It reduces the dependency on initialization and then reduces overfitting due to its minor regularization effect. 
Similarly to dropout, it adds some noise to each hidden layer’s activation.
\item Since the scale of input features would not differ significantly, the gradient descent may reduce the oscillations when approaching the optimum and thus converge faster.
\item BN reduces the impacts of earlier layers on the following layers in DNNs. 
Therefore, it takes more time to train the model to converge. However, the use of BN can reduce the impact of earlier layers by keeping the mean and variance fixed, which in some way makes the layers independent from each other. 
Consequently, the convergence becomes faster.
\end{itemize}
The development of BN as a normalization technique was a turning point in the development of DL models, and it enabled various networks to train and converge. Despite its great success, BN exhibits drawbacks that are caused by its distinct behavior of normalizing along the batch dimension. 
One of the major disadvantages of BN is that it requires sufficiently large batch sizes to obtain good results. 
This prevents the user from exploring higher-capacity models that would be limited by memory. 
To solve this problem, several other normalization variants are developed, such as Layer Normalization (LN)~\cite{ba2016layer}, Instance Normalization (IN)~\cite{ulyanov2016instance}, Group Normalization (GN)~\cite{wu2018group} and Filter Response Normalization (FRN)~\cite{singh2019filter}.
\section{Related Work}
\label{sec:relatedWork}
Investigating the effect of learnable parameters of BN, scale ($\gamma$) and shift ($\beta$), on the training of various typical deep neural nets, Wang et al.~\cite{wang2018batch} suggest that there is no big difference in both training convergence and final test accuracy when removing the BN layer following the final convolutional layer from a convolutional neural network (CNN) for standard classification tasks. The authors claim that it is not necessary to adjust the learnable gain and bias along the training process and they show that the use of constant gain and bias is enough and these learneable parameters have little effect on the performance. They also observe that without adaptively updating learnable parameters for BN layers, it often requires less time for training of very deep neural nets such as ResNet-101. 

Frankle et al.~\cite{frankle2020training} also studied the effect of learnable BN parameters, showing that BN’s trainable parameters alone can account for much of a network’s accuracy. They found that,
when locking all other layers at their random initial weights and then training the network for fifty or so epochs, it will perform better than random. As it adjusts a network’s intermediate feature representations for a given mini batch, BN itself learns how to do so in a consistent way for all mini batches. The researchers probed the impact of this learning by training only the BN parameters, $\gamma$ and $\beta$, while setting all other parameters at random.   

Bjorck et al.~\cite{bjorck2018understanding}
conducted another experiment, similar to our study. They trained a ResNet that uses one BN layer only at the very last convolutional layer of the network (removing all the other BN layers coming after each CNN layer), normalizing the output of the last residual block but without intermediate activation. This indicates that after initialization, the network tends to almost always predict the same (typically wrong) class, which is then corrected with a strong gradient update. In contrast, the network with BN does not exhibit the same behavior; rather, positive gradients are distributed throughout all classes. Their findings suggest that normalizing the final layer of a deep network may be one of the most important contributions of BN. 

Zhu et al.~\cite{zhu2020improving} also adopted a similar idea of using the scalable version of a BN layer to normalize the channel at the final output of the network, and improving the classification performance of softmax without adding learnable BN parameters, but controlling the output distribution by another scaling parameter. However, their improvements on selected datasets were not significant (less then 0.3\%).

Apart from these studies, to the best of our knowledge, there is no prior work investigating the effect of BN utilization over imbalanced classification tasks when set before the softmax output layer.

\section{Implementation Details and Experimental Results}
\label{sec:ImplementationDetails}
In this study, we primarily utilize the ResNet34 CNN architecture due to computational requirements and a need for an iterative process to conduct a large number of experiments. ResNet is evaluated on the ImageNet with a depth of up to 152 layers - 8 times deeper than VGG nets but still having lower complexity. An ensemble of these residual nets achieved a 3.57\% error on the ImageNet test set. This result won the 1st place in the ILSVRC 2015 classification task~\cite{he2016deep}. 

We firstly addressed the complete PlantVillage original dataset and trained a ResNet34 model for 38 classes. Using scheduled learning rates, we obtained 99.782\% accuracy after 10 epochs -- slightly improving the PlantVillage project's record of 99.34\% using GoogleNet~\cite{mohanty2016using}. 
Having reproducing these results, relatively easily, further boosts our confidence in selecting ResNet34 for the task.

\subsection{Adding a Final Batch Norm Layer Before the Output Layer}
By using the imbalanced datasets for certain plant types (1,000/10 in the training set, 150/7 in the validation set and 150/150 in the test set), we performed several experiments with the VGG19 and ResNet34 architectures.
The selected plant types were Apple, Pepper and Tomato - being the only datasets of sufficient size to enable the 99\%-1\% skewness generation.

\begin{figure}[h]
\centering
\includegraphics[width=0.4\textwidth,scale=0.4]{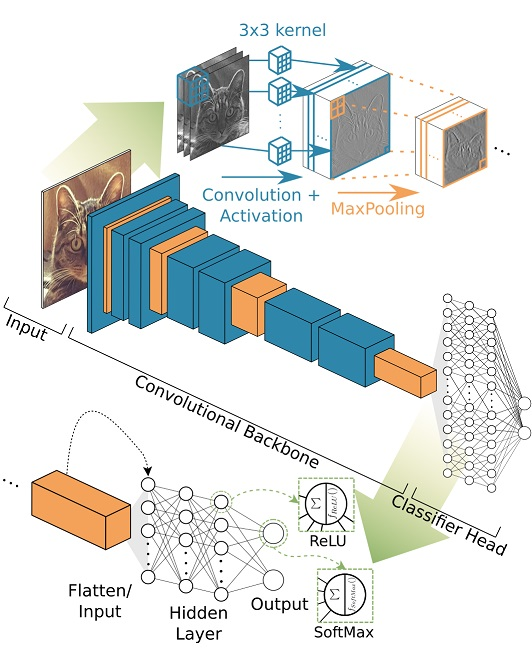}
\centering
\caption{Overview and details of a convolutional neural network (CNN) architecture for image recognition.~\cite{hoeser2020object}}
\label{fig:cnn_framework}
\end{figure}

\begin{figure*}[h]
\includegraphics[width=\textwidth,scale=1.3]{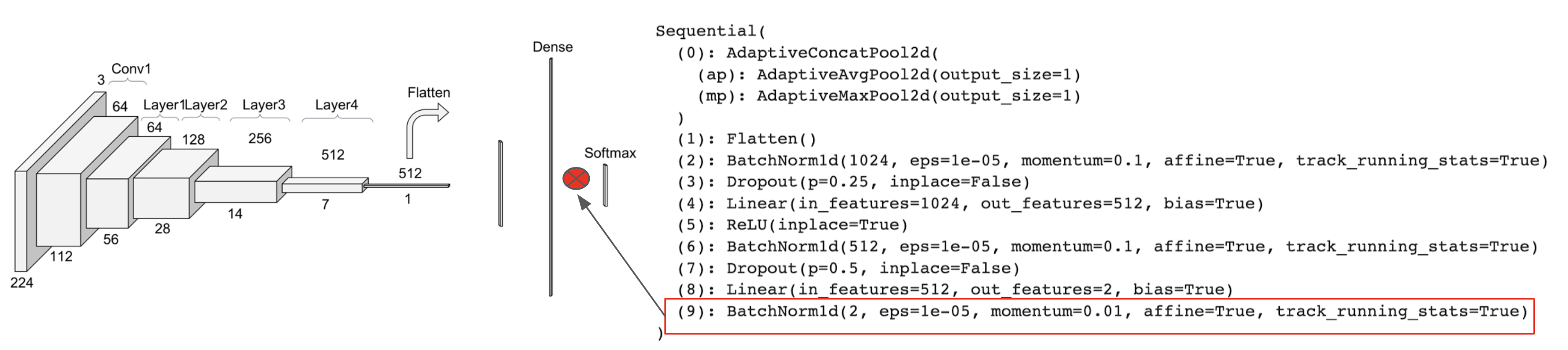}
\caption{Implementation of the final BN layer in ResNet34, just before the softmax output.}
\label{fig:BN_last_layer_code}
\end{figure*}

In order to fine-tune our network for the PlantVillage dataset, the final classification layer of CNN architectures is replaced by Adaptive Average Pooling (AAP), BN, Dropout, Dense, ReLU, BN and Dropout followed by the Dense and BN layer again. The last layer of an image classification network is often a fully-connected layer with a hidden size being equal to the number of labels to output the predicted confidence scores that are normalized by the softmax operator to obtain predicted probabilities. In our implementation, we add another 2-input BN layer after the last dense layer (before softmax output) in addition to existing BN layers in the tail and 4 BN layers in the head of the DL architecture (e.g., ResNet34 possesses a BN layer after each convolutional layer, having altogether 38 BN layers given the additional 4 in the head). A schematic outline of this architecture is provided in Figure~\ref{fig:BN_last_layer_code} and details of a typical CNN architecture for image recognition can be seen at Figure~\ref{fig:cnn_framework}.

\begin{figure*}[h]
\includegraphics[width=\textwidth]{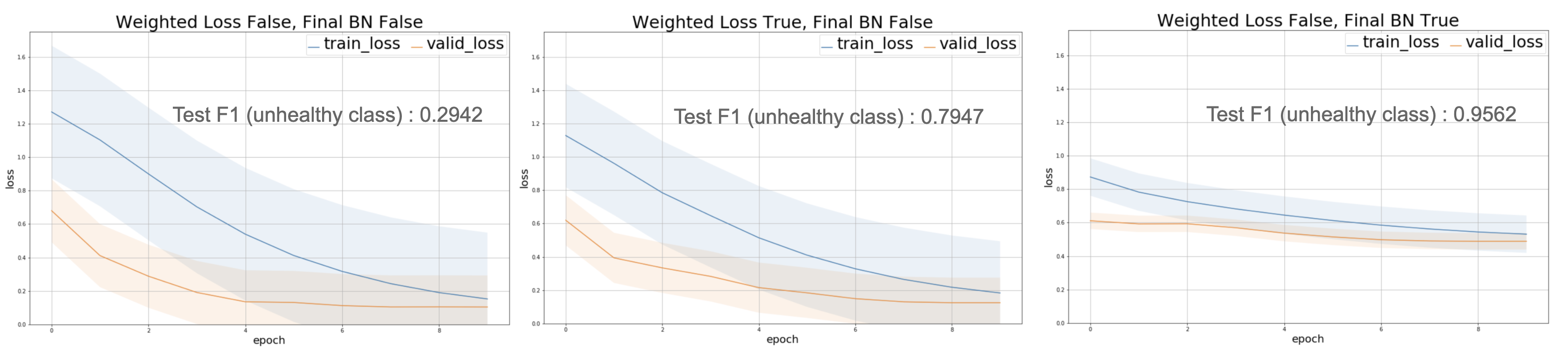}
\caption{Averaged loss charts over 10 runs for the Apple dataset without final BN (left), with final BN layer (right) and with WL (center) in VGG19 architecture. Without WL, after adding the final BN layer, the test F1 score is increased from 0.2942 to 0.9562 for the minority class while the train and validation losses level off around 0.5 before the model begins to overfit. 
It is also evident that the standard deviation among runs (depicted as the shaded areas surrounding a given curve within each plot) is lower for the latter case, as the training becomes smoother therein.}
\label{fig:apple_loss_3fig}
\end{figure*}

At first we run experiments with VGG19 architectures for selected plant types by adding the final BN layer. When we train this model for 10 epochs and repeat this for 10 times, we observed that the F1 test score is increased from 0.2942 to 0.9562 for unhealthy Apple, from 0.7237 to 0.9575 for unhealthy Pepper and from 0.5688 to 0.9786 for unhealthy Tomato leaves. We also achieved substantial improvement in healthy samples (being the majority in the training set). For detailed metrics and charts, see Figure~\ref{fig:apple_loss_3fig} and Table~\ref{tab:three_plants}.

\begin{table}[!htbp]
\caption{Averaged \textbf{F1 test set} performance values over 10 runs, alongside BN's total improvement, using 10 epochs with VGG19, with/without BN and with Weighted Loss (WL) without BN.
}
\label{tab:three_plants}
\resizebox{\columnwidth}{!}{
\begin{tabular}{llcccc}
\toprule
plant & class & \multicolumn{1}{c}{\thead{without\\ final BN}} & \thead{with WL\\ (no BN)} & \multicolumn{1}{c}{\thead{with final BN\\ (no WL)}} & \thead{BN total\\ improvement} \\
\midrule
Apple  & Unhealthy & \textbf{0.2942} & \textbf{0.7947} & \textbf{0.9562} & \textbf{0.1615}\\
       & Healthy   & 0.7075 & 0.8596 & 0.9577 & 0.0981\\ [0.05cm]
Pepper & Unhealthy & \textbf{0.7237} & \textbf{0.8939} & \textbf{0.9575} & \textbf{0.0636} \\ 
       & Healthy   & 0.8229 & 0.9121 & 0.9558 & 0.0437\\ [0.05cm]
Tomato & Unhealthy & \textbf{0.5688} & \textbf{0.8671} & \textbf{0.9786} & \textbf{0.1115}\\
       & Healthy   & 0.7708 & 0.9121 & 0.9780 & 0.0659\\
\bottomrule
\end{tabular}
}
\end{table}

\subsection{Experimentation subject to Different Configurations}
Using the following six configuration variations with two options each, we created 64 different configurations which we tested with ResNet34 (training for 10 epochs only):
Adding (\checkmark) a final BN layer just before the output layer (BN), using (\checkmark) weighted cross-entropy loss~\cite{goodfellow2016deep} according to class imbalance (WL), using (\checkmark) data augmentation (DA), using (\checkmark) mixup (MX)~\cite{zhang2017mixup}, unfreezing  (\checkmark) or freezing (learnable vs pre-trained weights) the previous BN layers in ResNet34 (UF), and using (\checkmark) weight decay (WD)~\cite{krogh1992simple}. 
Checkmarks (\checkmark) and two-letter abbreviations are used in Table~\ref{best_scores} and Table~\ref{best_worst_metrics} to denote configurations. When an option is disabled across all configurations, its associated column is dropped.
%

\begin{table*}[h]
\caption{Best performance metrics over the Apple dataset under various configurations using ResNet34.}
\label{best_scores}
\centering
\begin{tabular}{ccccccccccc}\toprule
\multicolumn{1}{l}{Class} &
  \thead{Config\\ Id} &
  \thead{Test set\\ precision} &
  \thead{Test set\\  recall} &
  \thead{Test set\\ F1-score} &
  Epoch &
    BN &
    DA &
    UF &
    WD \\
\midrule
Unhealthy & \textbf{31} & 0.9856 & 0.9133 & \textbf{0.9481} & 6 & \checkmark  & & &  \\ 
($\text{class}=1$) & 23 & 0.9718 & 0.9200 & 0.9452 & 6 & \checkmark  & \checkmark  & & \\ 
          & 20 & 0.9926 & 0.8933 & 0.9404 & 7 & \checkmark  & \checkmark  & \checkmark  & \checkmark \\ [0.1cm] 
Healthy   & 31 & 0.9193 & 0.9867 & 0.9518 & 6 & \checkmark  & & & \\ 
($\text{class}=0$) & 23 & 0.9241 & 0.9733 & 0.9481 & 6 & \checkmark  & \checkmark  & & \\ 
          & 20 & 0.9030 & 0.9933 & 0.9460 & 7 & \checkmark  & \checkmark  & \checkmark  & \checkmark \\ 
\bottomrule
\end{tabular}
\end{table*}

As shown in Table~\ref{best_scores}, just adding the final BN layer was enough to get the highest F1 test score in both classes. Surprisingly, although there is already a BN layer after each convolutional layer in the base CNN architecture, adding one more BN layer just before the output layer boosts the test scores. 
Notably, the 3rd best score (average score for configuration 31 in Table~\ref{best_worst_metrics}) 
is achieved just by adding a single BN layer before the output layer, even without unfreezing the previous BN layers. 
Moreover, it is evident that the additional BN layer is never utilized among the worst performing configurations (see Table~\ref{best_worst_metrics}).
%
%
%
%
%
\begin{table*}[htbp]
\caption{Best (top three) and worst (bottom three) performing configurations (F1 measure, for the unhealthy/minority class) when using ResNet34 for the Apple, Pepper and Tomato datasets.}
\label{best_worst_metrics}
\centering
\begin{tabular}{cccccccccc}
\toprule
Config Id & Apple & Pepper & Tomato & Average & BN  & DA & MX & UF & WD \\
\midrule
29 & 0.9332 & 0.9700 & 0.9866 & \textbf{0.9633} & \checkmark   &  &  & \checkmark & \\ 
28 & 0.9332 & 0.9566 & 0.9966 & 0.9622 & \checkmark   &  &  & \checkmark  & \checkmark  \\
\textbf{31} & 0.9499 & 0.9433 & 0.9833 & \textbf{0.9588} & \checkmark   &  &  &  &  \\ [0.1cm]
49 & 0.6804 & 0.5080 & 0.5339 & 0.5741 &  & \checkmark  & \checkmark  & \checkmark  &  \\
48 & 0.5288 & 0.5638 & 0.5638 & 0.5521 &  & \checkmark  & \checkmark  & \checkmark  & \checkmark  \\
50 & 0.6377 & 0.5027 & 0.4528 & 0.5311 &  & \checkmark  & \checkmark  &  & \checkmark  \\ 
\bottomrule
\end{tabular}
\end{table*}

The model without the final BN layer is pretty confident even if it predicts falsely. 
But the proposed model with the final BN layer predicts correctly even though it is less confident. We end up with less confident but more accurate models in less than 10 epochs. 
The classification probabilities for five sample images from the unhealthy class ($\text{class}=1$) with final BN layer (right column) and without final BN layer (left column) are shown in Table~\ref{tab:confident_wrongs}.
As explained above, without the final BN layer, these anomalies are all falsely classified (recall that ${\cal P}_{\text{softmax}}(\text{class} = 0) = 1 - {\cal P}_{\text{softmax}}(\text{class} = 1)$).
%

\begin{table}[htbp]
\caption{Softmax output values (representing class probabilities) for five sample images of unhealthy 
plants. Left column: Without final BN layer, softmax output values for unhealthy, resulting in a wrong classification in each case. Right column: With final BN layer, softmax output value for unhealthy, resulting in correct but less "confident" classifications.}
\label{tab:confident_wrongs}
\centering
\begin{tabular}{cc}
\toprule
\multicolumn{1}{c}{Without final BN layer} &
\multicolumn{1}{c}{With final BN layer} \\
\midrule
0.1082 & 0.5108\\
0.1464 & 0.6369\\
0.1999 & 0.6082\\
0.2725 & 0.6866\\
0.3338 & 0.7032\\
\bottomrule
\end{tabular}
\end{table}

\section{Discussion}
\label{sec:discussion}

To see what would have happened had we let the models further train, we conducted a 100-epoch training for both models and compared the results. We observed that the test accuracy with the model having a final BN layer quickly rises to 94\% within only 6 epochs, but then decreases steadily to 78\% within a total of 100 epochs.

%

\begin{figure*}[h]
\begin{center}
\includegraphics[width=0.9\textwidth]{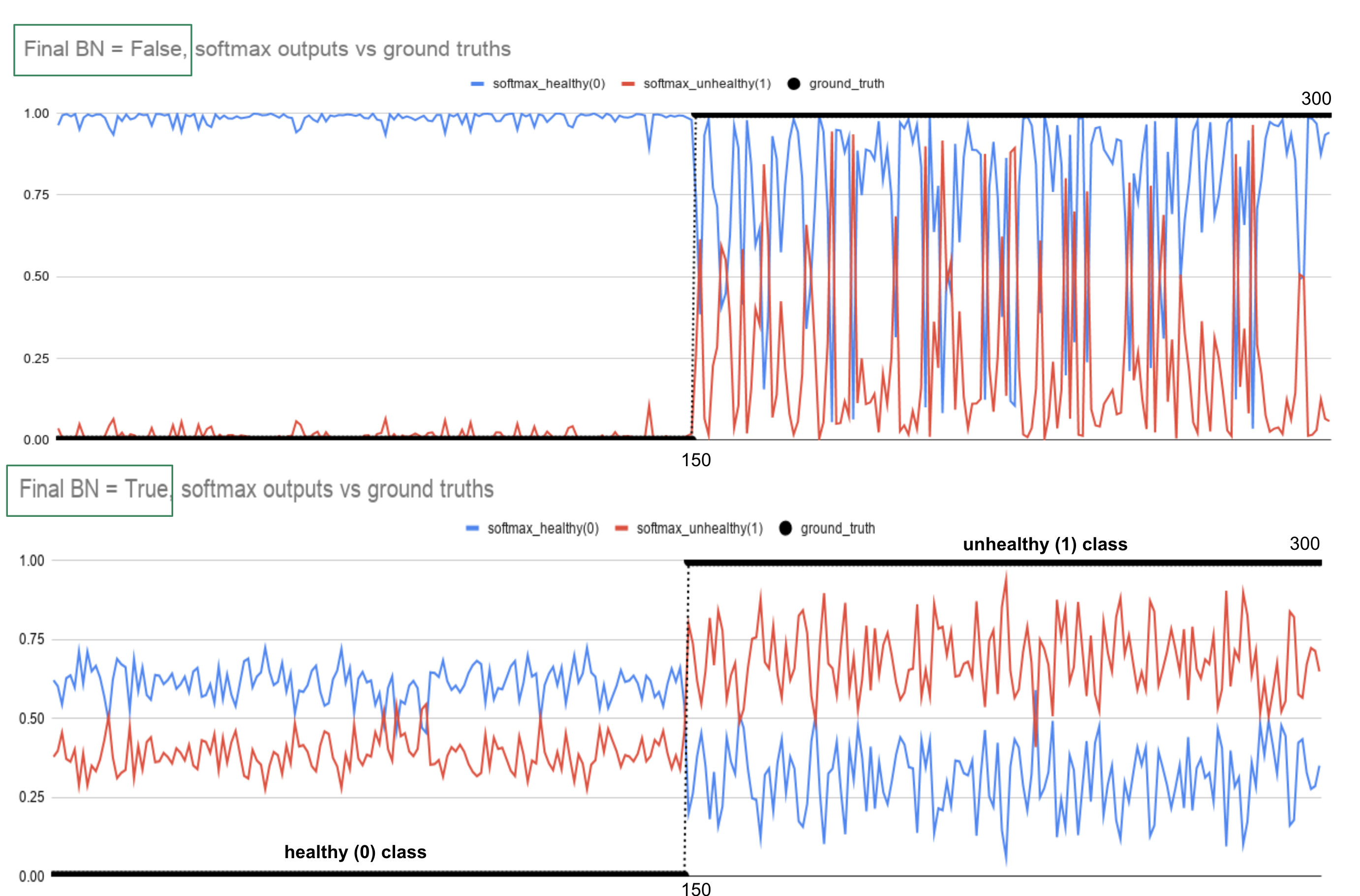}
\end{center}
\caption{The $x$-axis represents the ground truth for all 150 healthy (0) and 150 unhealthy (1) images in the test set while red and blue lines represent final softmax output values between 0 and 1 for each image. \textbf{Top chart (without final BN):} When ground truth (black) is $\text{class} = 0$ (healthy), the softmax output for $\text{class} = 0$ is around 1.0 (blue, predicting correctly). But when ground truth (black) is $\text{class} = 1$ (unhealthy), the softmax output for $\text{class} = 1$ (red points) changes between 0.0 and 1.0 (mostly below 0.5, NOT predicting correctly). \textbf{Bottom chart (with final BN):} When ground truth (black) is $\text{class} = 0$ (healthy), the softmax output is between 0.5 and 0.75 (blue, predicting correctly). When ground truth (black) is $\text{class} = 1$ (unhealthy), the softmax output  (red points) changes between 0.5 and 1.0 (mostly above 0.5, predicting correctly).}
\label{fig:softmax_outputs}
\end{figure*}

By adding a final BN layer just before the output, we normalize the output of the final dense layer, and feed into the softmax layer so that we shrink the gap between class activations (see  Figure~\ref{fig:softmax_outputs}). As its name suggests, softmax will always favor the large values when the layer outputs spread over large ranges. By applying BN to dense layer outputs, the gap between activations is reduced (normalized) and then softmax is applied on normalized outputs, which are centered around the mean. 
Therefore, we end up with centered probabilities (around 0.5), but favoring the minority class by a small margin. 


This shows that DNNs have the tendency of becoming ‘over-confident’ in their predictions during training, and this can reduce their ability to generalize and thus perform as well on unseen data. In addition, large datasets can often comprise incorrectly labeled data, meaning inherently the DNN should be a bit skeptical of the ‘correct answer’ to avoid being overconfident on bad answers. This was the main motivation of M{\"u}ller et al.~\cite{muller2019does} for proposing the \textit{label smoothing}, a loss function modification that has been shown to be effective for training DNNs. Label smoothing encourages the activations of the penultimate layer to be close to the template of the correct class and equally distant to the templates of the incorrect classes~\cite{muller2019does}. Despite its relevancy to our research, label smoothing did not do well in our study as previously mentioned by Kornblith et al.~\cite{chelombiev2019adaptive} who demonstrated that label smoothing impairs the accuracy of transfer learning, which similarly depends on the presence of non-class-relevant information in the final layers of the network.

Another observation is that a CNN with final BN is more calibrated when compared to an equivalent network lacking that layer. 
The calibration principle~\cite{guo2017calibration} states that \textit{the probability associated with the predicted class label should reflect its ground truth correctness likelihood}. A classification model is calibrated if its predicted outcome probabilities reflect their accuracy.
We tested the calibration of both networks (with and without final BN) by drawing the reliability diagrams as suggested by Guo et al.~\cite{guo2017calibration} and found that when the final BN layer is added, a network has much lower ECE (Expected Calibration Error), i.e., ideal confidence relative to its own accuracy. In effect, a final BN layered network is not ‘over-confident’ and as a result generalizes and performs better on real world data. This is also confirmed by the fact that the network with a final BN layer has a lower Brier score~\cite{brier1950verification} \textit{(squared error between the predicted probability vector and the one-hot encoded true response)} than without a final BN layer.
Figure~\ref{fig:softmax_outputs} shows that softmax outputs level off around 0.5, as also depicted in Figure~\ref{fig:apple_loss_3fig}. Since the network loss is optimized by using softmax output value and the ground truth, focusing too much on loss minimization may not be viable when the final BN layer is used in similar settings. 

\subsection*{Additional Experiments}
In addition to the aforementioned experiments, we ran more tests to further support our findings:
Turning off the bias at the final dense layer when the final BN layer is added, using SeLU~\cite{klambauer2017self} instead of BN in the last layer, freezing and unfreezing the previous BN layers, label smoothing, training a simple CNN from scratch rather than utilizing SOTA architectures. 

Also, additional corroboration was achieved over the ISIC Skin Cancer dataset~\cite{gutman2016skin} as well as the Wall Crack dataset~\cite{zhang2016road} exhibiting similar patterns of behavior and thus supporting our reported findings.
These observations and the code to reproduce the results mentioned in this paper are fully reported in the supplementary material. 

\section{Conclusions}
\label{sec:conclusion}
This study illustrates that putting an additional BN layer just before the output layer has a considerable impact in terms of minimizing the training time and test error for minority classes in highly imbalanced datasets. 
As evident from Table~\ref{tab:three_plants}, upon adding the final BN layer the F1 test score is increased from 0.2942 to 0.9562 for the unhealthy Apple minority class, from 0.7237 to 0.9575 for the unhealthy Pepper and from 0.5688 to 0.9786 for the unhealthy Tomato when WL is not used (all are averaged values over 10 runs).  When compared to a network with WL but without a final BN layer, the averaged improvements were 0.1615, 0.0636 and 0.1115 respectively.
We show that the highest gain in test F1 score for both classes (majority vs.~minority) is achieved just by adding a final BN layer, resulting in a more than three-fold performance boost on some configurations.

We also argue that trying to minimize validation and train losses may not be an optimal way of getting a high F1 test score for minority classes. We presented that having a higher train and validation loss but high validation accuracy would lead to higher F1 test scores for minority classes in less time. That is, the model might perform better even if it is not confident enough while making a prediction. So, having a confident model (lower loss, higher accuracy) may favor the majority class in the short run and ends up with lower F1 test scores for minority class due to the conjecture that it learns to minimize error in majority class regardless of class balance.
Since BN (like dropout) adds stochasticity to the network, and the network learns to be robust to this stochasticity during the entire training, it was expected that the final BN layer would be detrimental for the network to cope with the stochasticity right before the output~\cite{zhu2020improving}.
However, we observed the contrary in our experiments. 

This study also shows that lower values in the softmax output may not necessarily indicate ‘lower confidence level’, leading to another discussion why softmax output may not serve as a good uncertainty measure for DNNs. In classification, predictive probabilities obtained at the end of the pipeline (the softmax output) are often wrongly interpreted as model confidence. As we have clearly shown, a model can be uncertain in its predictions even when having a high softmax output~\cite{gal2016dropout}.


\bibliographystyle{unsrtnat}
\bibliography{References}

\begin{thebibliography}{35}
\providecommand{\natexlab}[1]{#1}
\providecommand{\url}[1]{\texttt{#1}}
\expandafter\ifx\csname urlstyle\endcsname\relax
  \providecommand{\doi}[1]{doi: #1}\else
  \providecommand{\doi}{doi: \begingroup \urlstyle{rm}\Url}\fi

\bibitem[Mahlein et~al.(2012)Mahlein, Oerke, Steiner, and
  Dehne]{mahlein_recent_2012}
Anne-Katrin Mahlein, Erich-Christian Oerke, Ulrike Steiner, and Heinz-Wilhelm
  Dehne.
\newblock Recent advances in sensing plant diseases for precision crop
  protection.
\newblock \emph{European Journal of Plant Pathology}, 133\penalty0
  (1):\penalty0 197--209, 2012.
\newblock ISSN 0929-1873, 1573-8469.
\newblock \doi{10.1007/s10658-011-9878-z}.
\newblock URL \url{http://link.springer.com/10.1007/s10658-011-9878-z}.

\bibitem[Hwang et~al.(2014)Hwang, Wijekoon, Kalischuk, Johnson, Howard,
  Pr{\"u}fer, and Kawchuk]{PotatoLateBlightReview2014}
Yeen~Ting Hwang, Champa Wijekoon, Melanie Kalischuk, Dan Johnson, Ron Howard,
  Dirk Pr{\"u}fer, and Lawrence Kawchuk.
\newblock Evolution and management of the irish potato famine pathogen
  phytophthora infestans in canada and the united states.
\newblock \emph{American Journal of Potato Research}, 91, 2014.

\bibitem[Duarte-Carvajalino et~al.(2018)Duarte-Carvajalino, Alzate, Ramirez,
  Santa-Sepulveda, Fajardo-Rojas, and Soto-Su{\'a}rez]{Monsanto2018}
Julio~M. Duarte-Carvajalino, Diego~F. Alzate, Andr{\'e}s~A. Ramirez, Juan~D.
  Santa-Sepulveda, Alexandra~E. Fajardo-Rojas, and Mauricio Soto-Su{\'a}rez.
\newblock Evaluating late blight severity in potato crops using unmanned aerial
  vehicles and machine learning algorithms.
\newblock \emph{Remote Sensing}, 10, 2018.

\bibitem[Beggel et~al.(2019)Beggel, Pfeiffer, and Bischl]{beggel2019robust}
Laura Beggel, Michael Pfeiffer, and Bernd Bischl.
\newblock Robust anomaly detection in images using adversarial autoencoders.
\newblock \emph{arXiv preprint arXiv:1901.06355}, 2019.

\bibitem[Hussain et~al.(2018)Hussain, Bird, and Faria]{hussain2018study}
Mahbub Hussain, Jordan~J Bird, and Diego~R Faria.
\newblock A study on {CNN} transfer learning for image classification.
\newblock In \emph{UK Workshop on Computational Intelligence}, pages 191--202.
  Springer, 2018.

\bibitem[Shorten and Khoshgoftaar(2019)]{shorten2019survey}
Connor Shorten and Taghi~M Khoshgoftaar.
\newblock A survey on image data augmentation for deep learning.
\newblock \emph{Journal of Big Data}, 6\penalty0 (1):\penalty0 60, 2019.

\bibitem[Barz and Denzler(2020)]{barz2020deep}
Bjorn Barz and Joachim Denzler.
\newblock Deep learning on small datasets without pre-training using cosine
  loss.
\newblock In \emph{The IEEE Winter Conference on Applications of Computer
  Vision}, pages 1371--1380, 2020.

\bibitem[Lake et~al.(2015)Lake, Salakhutdinov, and Tenenbaum]{lake2015human}
Brenden~M Lake, Ruslan Salakhutdinov, and Joshua~B Tenenbaum.
\newblock Human-level concept learning through probabilistic program induction.
\newblock \emph{Science}, 350\penalty0 (6266):\penalty0 1332--1338, 2015.

\bibitem[Hughes et~al.(2015)Hughes, Salath{\'e}, et~al.]{hughes2015open}
David Hughes, Marcel Salath{\'e}, et~al.
\newblock An open access repository of images on plant health to enable the
  development of mobile disease diagnostics.
\newblock \emph{arXiv preprint arXiv:1511.08060}, 2015.

\bibitem[Mohanty et~al.(2016)Mohanty, Hughes, and
  Salath{\'e}]{mohanty2016using}
Sharada~P Mohanty, David~P Hughes, and Marcel Salath{\'e}.
\newblock Using deep learning for image-based plant disease detection.
\newblock \emph{Frontiers in plant science}, 7:\penalty0 1419, 2016.

\bibitem[He et~al.(2016)He, Zhang, Ren, and Sun]{he2016deep}
Kaiming He, Xiangyu Zhang, Shaoqing Ren, and Jian Sun.
\newblock Deep residual learning for image recognition.
\newblock In \emph{Proceedings of the IEEE conference on computer vision and
  pattern recognition}, pages 770--778, 2016.

\bibitem[Simonyan and Zisserman(2014)]{simonyan2014very}
Karen Simonyan and Andrew Zisserman.
\newblock Very deep convolutional networks for large-scale image recognition.
\newblock \emph{arXiv preprint arXiv:1409.1556}, 2014.

\bibitem[Gutman et~al.(2016)Gutman, Codella, Celebi, Helba, Marchetti, Mishra,
  and Halpern]{gutman2016skin}
David Gutman, Noel~CF Codella, Emre Celebi, Brian Helba, Michael Marchetti,
  Nabin Mishra, and Allan Halpern.
\newblock Skin lesion analysis toward melanoma detection: A challenge at the
  international symposium on biomedical imaging (isbi) 2016, hosted by the
  international skin imaging collaboration (isic).
\newblock \emph{arXiv preprint arXiv:1605.01397}, 2016.

\bibitem[Zhang et~al.(2016)Zhang, Yang, Zhang, and Zhu]{zhang2016road}
Lei Zhang, Fan Yang, Yimin~Daniel Zhang, and Ying~Julie Zhu.
\newblock Road crack detection using deep convolutional neural network.
\newblock In \emph{2016 IEEE international conference on image processing
  (ICIP)}, pages 3708--3712. IEEE, 2016.

\bibitem[Ioffe and Szegedy(2015)]{ioffe2015batch}
Sergey Ioffe and Christian Szegedy.
\newblock Batch normalization: Accelerating deep network training by reducing
  internal covariate shift.
\newblock \emph{arXiv preprint arXiv:1502.03167}, 2015.

\bibitem[Santurkar et~al.(2018)Santurkar, Tsipras, Ilyas, and
  Madry]{santurkar2018does}
Shibani Santurkar, Dimitris Tsipras, Andrew Ilyas, and Aleksander Madry.
\newblock How does batch normalization help optimization?
\newblock In \emph{Advances in Neural Information Processing Systems}, pages
  2483--2493, 2018.

\bibitem[Bjorck et~al.(2018)Bjorck, Gomes, Selman, and
  Weinberger]{bjorck2018understanding}
Nils Bjorck, Carla~P Gomes, Bart Selman, and Kilian~Q Weinberger.
\newblock Understanding batch normalization.
\newblock In \emph{Advances in Neural Information Processing Systems}, pages
  7694--7705, 2018.

\bibitem[Mishkin and Matas(2015)]{mishkin2015all}
Dmytro Mishkin and Jiri Matas.
\newblock All you need is a good init.
\newblock \emph{arXiv preprint arXiv:1511.06422}, 2015.

\bibitem[Ba et~al.(2016)Ba, Kiros, and Hinton]{ba2016layer}
Jimmy~Lei Ba, Jamie~Ryan Kiros, and Geoffrey~E Hinton.
\newblock Layer normalization.
\newblock \emph{arXiv preprint arXiv:1607.06450}, 2016.

\bibitem[Ulyanov et~al.(2016)Ulyanov, Vedaldi, and
  Lempitsky]{ulyanov2016instance}
Dmitry Ulyanov, Andrea Vedaldi, and Victor Lempitsky.
\newblock Instance normalization: The missing ingredient for fast stylization.
\newblock \emph{arXiv preprint arXiv:1607.08022}, 2016.

\bibitem[Wu and He(2018)]{wu2018group}
Yuxin Wu and Kaiming He.
\newblock Group normalization.
\newblock In \emph{Proceedings of the European Conference on Computer Vision
  (ECCV)}, pages 3--19, 2018.

\bibitem[Singh and Krishnan(2019)]{singh2019filter}
Saurabh Singh and Shankar Krishnan.
\newblock Filter response normalization layer: Eliminating batch dependence in
  the training of deep neural networks.
\newblock \emph{arXiv preprint arXiv:1911.09737}, 2019.

\bibitem[Wang et~al.(2018)Wang, Wu, Chang, Zhang, Zhou, and Yan]{wang2018batch}
Yan Wang, Xiaofu Wu, Yuanyuan Chang, Suofei Zhang, Quan Zhou, and Jun Yan.
\newblock Batch normalization: Is learning an adaptive gain and bias necessary?
\newblock In \emph{Proceedings of the 2018 10th International Conference on
  Machine Learning and Computing}, pages 36--40, 2018.

\bibitem[Frankle et~al.(2020)Frankle, Schwab, and Morcos]{frankle2020training}
Jonathan Frankle, David~J Schwab, and Ari~S Morcos.
\newblock Training batchnorm and only batchnorm: On the expressive power of
  random features in {CNN}s.
\newblock \emph{arXiv preprint arXiv:2003.00152}, 2020.

\bibitem[Zhu et~al.(2020)Zhu, He, Zhang, and Cui]{zhu2020improving}
Qiuyu Zhu, Zikuang He, Tao Zhang, and Wennan Cui.
\newblock Improving classification performance of softmax loss function based
  on scalable batch-normalization.
\newblock \emph{Applied Sciences}, 10\penalty0 (8):\penalty0 2950, 2020.

\bibitem[Hoeser and Kuenzer(2020)]{hoeser2020object}
Thorsten Hoeser and Claudia Kuenzer.
\newblock Object detection and image segmentation with deep learning on earth
  observation data: A review-part i: Evolution and recent trends.
\newblock \emph{Remote Sensing}, 12\penalty0 (10):\penalty0 1667, 2020.

\bibitem[Goodfellow et~al.(2016)Goodfellow, Bengio, and
  Courville]{goodfellow2016deep}
Ian Goodfellow, Yoshua Bengio, and Aaron Courville.
\newblock \emph{Deep learning}.
\newblock MIT press, 2016.

\bibitem[Zhang et~al.(2017)Zhang, Cisse, Dauphin, and
  Lopez-Paz]{zhang2017mixup}
Hongyi Zhang, Moustapha Cisse, Yann~N Dauphin, and David Lopez-Paz.
\newblock mixup: Beyond empirical risk minimization.
\newblock \emph{arXiv preprint arXiv:1710.09412}, 2017.

\bibitem[Krogh and Hertz(1992)]{krogh1992simple}
Anders Krogh and John~A Hertz.
\newblock A simple weight decay can improve generalization.
\newblock In \emph{Advances in neural information processing systems}, pages
  950--957, 1992.

\bibitem[M{\"u}ller et~al.(2019)M{\"u}ller, Kornblith, and
  Hinton]{muller2019does}
Rafael M{\"u}ller, Simon Kornblith, and Geoffrey~E Hinton.
\newblock When does label smoothing help?
\newblock In \emph{Advances in Neural Information Processing Systems}, pages
  4696--4705, 2019.

\bibitem[Chelombiev et~al.(2019)Chelombiev, Houghton, and
  O'Donnell]{chelombiev2019adaptive}
Ivan Chelombiev, Conor Houghton, and Cian O'Donnell.
\newblock Adaptive estimators show information compression in deep neural
  networks.
\newblock \emph{arXiv preprint arXiv:1902.09037}, 2019.

\bibitem[Guo et~al.(2017)Guo, Pleiss, Sun, and Weinberger]{guo2017calibration}
Chuan Guo, Geoff Pleiss, Yu~Sun, and Kilian~Q Weinberger.
\newblock On calibration of modern neural networks.
\newblock In \emph{Proceedings of the 34th International Conference on Machine
  Learning-Volume 70}, pages 1321--1330. JMLR. org, 2017.

\bibitem[Brier(1950)]{brier1950verification}
Glenn~W Brier.
\newblock Verification of forecasts expressed in terms of probability.
\newblock \emph{Monthly weather review}, 78\penalty0 (1):\penalty0 1--3, 1950.

\bibitem[Klambauer et~al.(2017)Klambauer, Unterthiner, Mayr, and
  Hochreiter]{klambauer2017self}
G{\"u}nter Klambauer, Thomas Unterthiner, Andreas Mayr, and Sepp Hochreiter.
\newblock Self-normalizing neural networks.
\newblock In \emph{Advances in neural information processing systems}, pages
  971--980, 2017.

\bibitem[Gal and Ghahramani(2016)]{gal2016dropout}
Yarin Gal and Zoubin Ghahramani.
\newblock Dropout as a bayesian approximation: Representing model uncertainty
  in deep learning.
\newblock In \emph{international conference on machine learning}, pages
  1050--1059, 2016.

\end{thebibliography}


\end{document}